\documentclass[letterpaper, 10 pt, conference]{ieeeconf}  %

\IEEEoverridecommandlockouts
\usepackage{amsmath}
\DeclareMathOperator*{\argmin}{arg\,min}
\usepackage[per-mode=fraction]{siunitx}
\usepackage[pdftex]{graphicx}
\usepackage{svg}
\graphicspath{{figures/}}
\svgpath{figures/}
\usepackage{tikz}
\usetikzlibrary{patterns, arrows, shapes, positioning}
\usepackage{pgfplots}
\pgfplotsset{compat=newest}
\usepackage{hyperref}
\hypersetup{
    colorlinks=true,
    linkcolor=black,
    citecolor=black,
    filecolor=black,
    urlcolor=black,
}
\usepackage[T1]{fontenc}
\usepackage[utf8]{inputenc}
\usepackage{csquotes}
\usepackage[english]{babel}
\usepackage[export]{adjustbox}
\usepackage{caption}
\usepackage{subcaption}
\usepackage[colorinlistoftodos]{todonotes}
\usepackage{lipsum}
\usepackage{multirow}
\usepackage{textcomp}
\usepackage{placeins}%

\usepackage[style=ieee,
            doi=false,
            url=false,
            mincitenames=1,
            maxcitenames=1,
            minbibnames=6,
            maxbibnames=6,
            backend=biber]{biblatex}  %
\AtBeginBibliography{\footnotesize}

\title{%
HD Map Generation from Noisy Multi-Route Vehicle Fleet Data on Highways with Expectation Maximization
}

\author{
    Fabian Immel$^{1}$,
    Richard Fehler$^{1}$,
    Mohammad M. Ghanaat$^{2}$, 
    Florian Ries$^{2}$, \\
    Martin Haueis$^{2}$ and 
    Christoph Stiller$^{3}$%
    \thanks{
        $^{1}$FZI Research Center for Information Technology, Karlsruhe, Germany
        {\tt\small \{immel, fehler\}@fzi.de}
    }%
    \thanks{
        $^{2}$Mercedes-Benz AG, Research \& Development, Sindelfingen, Germany
        {\tt\small \{mohammad.m.ghanaat, florian.ries, martin.haueis\}@mercedes-benz.com}
    }%
    \thanks{
        $^{3}$Institute of Measurement and Control Systems, Karlsruhe Institute of Technology (KIT),
        Karlsruhe, Germany
        {\tt\small stiller@kit.edu}
    }%
}

\newcommand\copyrighttext{%
  \footnotesize \textcopyright 2023 IEEE. Personal use of this material is permitted.
  Permission from IEEE must be obtained for all other uses, in any current or future
  media, including reprinting/republishing this material for advertising or promotional
  purposes, creating new collective works, for resale or redistribution to servers or
  lists, or reuse of any copyrighted component of this work in other works.}
\newcommand\copyrightnotice{%
\begin{tikzpicture}[remember picture,overlay]
\node[anchor=south,yshift=10pt] at (current page.south) {\fbox{\parbox{\dimexpr\textwidth-\fboxsep-\fboxrule\relax}{\copyrighttext}}};
\end{tikzpicture}%
}

\begin{document}
\maketitle
\copyrightnotice

\pagestyle{empty}

\begin{abstract}

    High Definition (HD) maps are necessary for many applications of automated driving (AD), but their manual creation and maintenance is very costly.
    Vehicle fleet data from series production vehicles can be used to automatically generate HD maps, but the data is often incomplete and noisy. We propose a system for the generation of HD maps from vehicle fleet data, which is tolerant to missing or misclassified detections and can handle drives with multiple routes, generating a single complete map, model-free and without prior reference lines. Using randomly selected drives as pivot drives, a step-wise lateral sampling of detections is performed. These sampled points are then clustered and aligned using Expectation Maximization (EM), estimating a lateral offset for each drive to compensate localization errors. The clustered points are replaced with the maxima of their probability density function (PDF) and connected to form polylines using a modified rectangular linear assignment algorithm. The data from vehicles on varying routes is then fused into a hierarchical singular map graph. The proposed approach achieves an average accuracy below 0.5 meters compared to a hand annotated ground truth map, as well as correctly resolving lane splits and merges, proving the feasibility of the use of vehicle fleet data for the generation of highway HD maps.


\end{abstract}
\section{Introduction}
\label{introduction}

High definition (HD) maps are an integral part of many applications of automated driving \cite{Ziegler2014}. %
Compared to standard definition (SD) maps, HD maps contain more detailed and accurate information about a road, such as detailed and centimeter-level accurate road boundaries and lane markings.

A large obstacle in the widespread adoption of automated driving is the cumbersome and costly generation and updating of HD maps, as normally special vehicles with a variety of sensors, such as accurate GNSS, LiDAR and cameras are used.
Creating and maintaining a HD map for a large highway network solely with specialized mapping vehicles requires considerable effort and expenses.

\begin{figure}[htb]
    \includegraphics[width=\columnwidth]{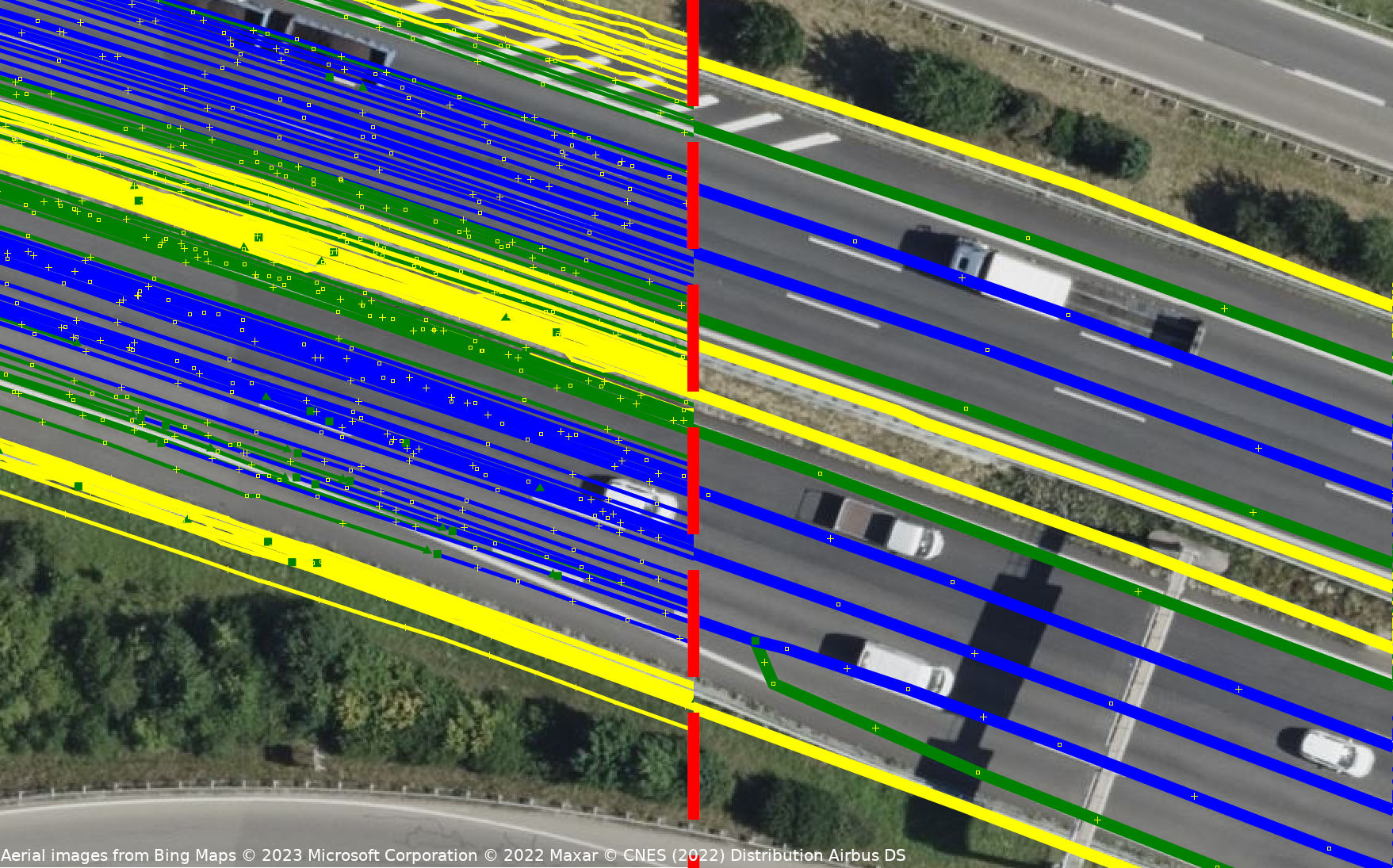}
    \caption{Result of the mapping (right) on a German highway overlaid with the input data (left). Displayed are the road barriers (yellow), dashed lines (blue) and solid lines (green).}
    \label{fig:sample_result}
    \vspace{-2mm}
\end{figure}

Due to the widespread proliferation of driver assistance systems in current series cars, a large body of lane marking and road boundary data is available.
Generating HD maps of a highway network from this data can enable a variety of systems and drastically reduce the costs of maintaining a HD map, as updates can be generated automatically from new fleet data.

However, processing this vehicle fleet data comes with several issues and requires aggregation and fusion of the individual detections.
Furthermore, the accuracy and completeness of the detected features is not on the same level as data from specialized mapping vehicles.
Inaccurate localization from series-level GNSS systems and missing detections of road features, possibly during large parts of a drive, complicate this aggregation and fusion.

\begin{figure*}[htb]
    \includegraphics[width=\linewidth]{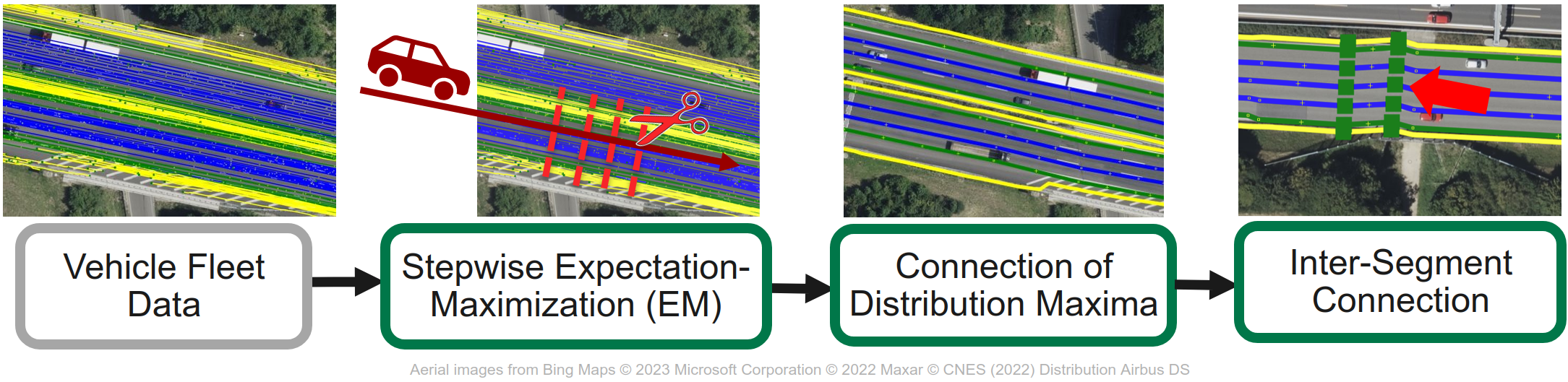}
    \caption{Flowchart with an overview of the map generation process.
        The input data is described in Sec. \ref{sec:input_data}, the stepwise EM in Sec. \ref{sec:stepwise_expectation_maximization}, the connection of distribution maxima in Sec. \ref{sec:connection_of_distribution_maxima} and the inter-segment connection in Sec. \ref{sec:inter_segment_connection}.
        The red arrow indicates the connection points of the inter-segment connection.}
    \label{fig:flowchart_overview}
    \vspace{-2mm}
\end{figure*}

In this paper we propose and evaluate a framework for a HD mapping pipeline, with an example of its results shown in Fig. \ref{fig:sample_result}.
It is capable of ingesting vehicle fleet data consisting of incomplete, partially misclassified and inaccurate detected lane markings and road boundaries and generating an HD map of the physical road features as connected polylines as a result.

Figure \ref{fig:flowchart_overview} shows an overview of the proposed map generation process.
The input data, connected lane marking detections of individual drives (Sec. \ref{sec:input_data}) are processed along randomly selected pivot drives by sampling detected points along a lateral intersection line (Sec. \ref{sec:lateral_sampling_of_detection_points}).
Sampled points along the ego-road are clustered using k-means and aligned by estimating a localization offset for each drive, being iteratively refined using EM (Sec. \ref{sec:expectation_maximization}).
The aggregated points are replaced with the maxima of their PDF (Sec. \ref{sec:aggregation_of_clustered_detection_points}) and connected to form a polyline (Sec. \ref{sec:connection_of_distribution_maxima}).
For connecting the points of consecutive sampling steps, a modified version of the Hungarian algorithm is used to find the optimal connection between all PDF maxima of the two sets. The thus modified connection algorithm is able to handle lane splits and merges.
Pivot drives visiting the same road segment along their route are skipped, reducing the computation time.

The data of different connected segments, e.g. from vehicles on varying routes, is then integrated into a hierarchical singular map graph (Sec. \ref{sec:inter_segment_connection}).

The main contributions of this work can be summarized as follows:
\begin{itemize}
    \item A novel end-to-end HD map generation pipeline capable of handling vehicle fleet data with incomplete, partially misclassified and inaccurate road feature detections.
    \item A model-free approach for the aggregation and fusion of detected lane markings and road boundaries, without prior reference lines.
    \item A graph-based method for the connection of fused detections capable of lane splits and merges and integration of multiple connected segments of different routes into a hierarchical singular map graph.
\end{itemize}
\section{Related Work}
\label{sec:Related-Work}

The automated generation of maps from crowdsourced or fleet data is a topic with longstanding research interest.

A large body of work focuses on the generation of SD maps and road networks, using driven trajectories as input data. Many approaches can be grouped into three categories \cite{Biagioni2012}: k-Means based \cite{Edelkamp2003, Schroedl2004}, Kernel Density Estimation (KDE) based \cite{Davies2006, Shi2009} and trace merge based \cite{Cao2009, Niehoefer2009}.
One commonality of all these approaches is that they only focus on basic road geometries similar to SD maps, since data about physical road features is not available.

With the recent increase in advanced driver assistance systems in vehicles, both the interest in HD maps and the availability of fleet data on physical road features has increased. This lead to a variety of research utilizing multi-journey data for the generation of HD maps.
Vision methods such as \cite{8968020, Dabeer2017} assume access to the recorded video data of the vehicles.
\cite{8968020} uses single camera images for visual SLAM to generate 3D boundaries of road feature elements.
\cite{Dabeer2017} describes an system to detect road features in single drives and then fuse these drives in a bundle adjustment pipeline.

Conceptually closer to the approach presented in this paper are \cite{DerivingHDMapsforHighlyAutomatedDrivingfromVehicularProbeData, HDMapGenerationfromVehicleFleetDataforHighlyAutomatedDrivingonHighways, Liebner2019}, as they use already processed road boundary and lane marking detections as their input.
\cite{DerivingHDMapsforHighlyAutomatedDrivingfromVehicularProbeData} corrects detected lane borders by grouping and aggregating detected left road boundaries and then fuses the corrected detections with k-means clustering.
\cite{HDMapGenerationfromVehicleFleetDataforHighlyAutomatedDrivingonHighways} uses unconnected road feature point measurements and drive trajectories to formulate a graph SLAM and aggregate the connections.
\cite{Liebner2019} also uses a graph slam, but fuses the individual detections using a road model that is selected with a beam-style search.

Common to the described systems for already processed detections is the focus on simple highway sections with mostly unchanging lanes.
Other requirements are the use of external reference lines \cite{HDMapGenerationfromVehicleFleetDataforHighlyAutomatedDrivingonHighways}, always available detections of certain features \cite{DerivingHDMapsforHighlyAutomatedDrivingfromVehicularProbeData} or only a single link sequence \cite{Liebner2019}.
Important issues not tackled yet in literature are also data sets with multiple routes and the connection of generated patches into a complete map. %

\section{Input Data}
\label{sec:input_data}

The algorithm takes as input a set of drives from a vehicle fleet, containing the drive trajectories and the associated detected road features.
Importantly, the drives may belong to different routes in a highway network and can have sparse or partially misclassified road feature detections as well as only series-grade localization with the respective large errors.
Detected points are assumed to be already connected to polylines if they belong to the same road feature, however these polylines do not need to be complete and may have gaps in their connections as well as misclassifications.
For preprocessing, detection polylines with a length below 3 meters are removed, as these are very likely misdetections and the number of points in remaining polylines is reduced using the Ramer-Douglas-Peucker algorithm.

The resulting detection polylines are then inserted into an axis-aligned bounding box tree \cite{Hart2020}, a data structure similar to kd-trees for use with shapes instead of single points.
We take advantage of this data structure to efficiently find detection polylines during \ref{sec:lateral_sampling_of_detection_points}.

\section{Stepwise Expectation Maximization}
\label{sec:stepwise_expectation_maximization}

\subsection{Step-wise Lateral Sampling of Detection Points}
\label{sec:lateral_sampling_of_detection_points}

\begin{figure}[htb]
    \includegraphics[width=\columnwidth]{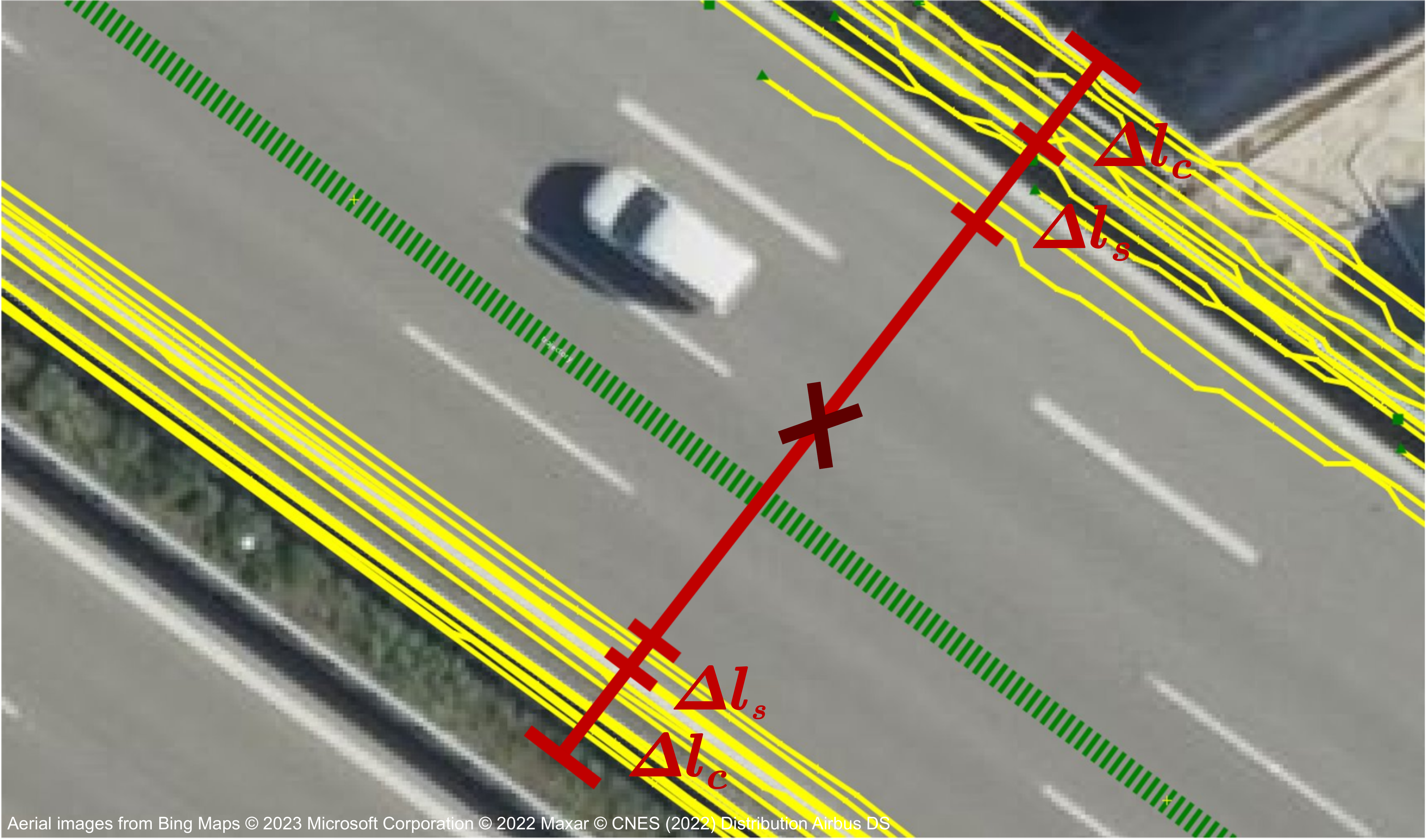}
    \caption{Visualization of the refined cut line calculation. The trajectory of the pivot drive can be seen in green, the filtered road boundary detections in yellow and the cut line in red. The line width is computed from the distance to the closest road boundary detections plus a margin $\Delta l_s$ depending on the detection spread and a constant margin $\Delta l_c$. The recomputed cut line midpoint is marked by the cross.}
    \label{fig:cut_line_visual}
    \vspace{-2mm}
\end{figure}

To avoid the requirement of reference lines that provide the basic road shape, the processing of detections is performed along pivot drives.
A pivot drive is randomly selected from the set of drives until all drives are processed and detections are sampled along its trajectory with small fixed-size steps in driving direction (e.g. 2 m).
Starting with the trajectory point as the midpoint, a lateral cut line segment is formed, whose intersection points with the detection polylines are sampled.
The purpose of the pivot drive is only to provide a very rough reference for the course of a road and to enable the lateral sampling of detections.
Sampled detections from drives from the opposite road direction are removed based on the heading information of the respective trajectories.

The goal of the lateral sampling is to obtain a set of all detections that belong to the ego-road of the pivot drive.
The road width is however a priori unknown and also cannot be inferred from the incomplete pivot drive detections.
Therefore the initial cut line segment has a fixed length of several times the average highway road width to ensure that all associated detections are included.
Since drives of various different routes are expected to be part of the data, refinement of the initial cut line segment and therefore filtering of the sampled detection points is necessary, to ensure that only detections belonging to the ego-road are included.

This refinement is performed with the help of the sampled detections, a visualization of which is shown in Fig. \ref{fig:cut_line_visual}.
A new cut line is calculated from the first road boundaries encountered from the original midpoint in both lateral directions, with the start and end point being extended by $\Delta l_s$ depending on the spread of the first road boundary detections encountered.
This provides a first estimate of the ego-road width, additionally including a constant margin $\Delta l_c$ so noisy detections are still included.
Length and midpoint of the new cut line segment are smoothed using a moving average filter with decreasing weights for older values and a regular moving average filter, respectively, to provide a road width estimate even in sections with missing road boundary detections.
The detections are then again sampled using the new cut line.

\subsection{Lateral Offset Estimation with Expectation Maximization}
\label{sec:expectation_maximization}

\begin{figure*}[!htb]
    \begin{subfigure}[c]{\columnwidth}
        \centering
        \includegraphics[width=\columnwidth]{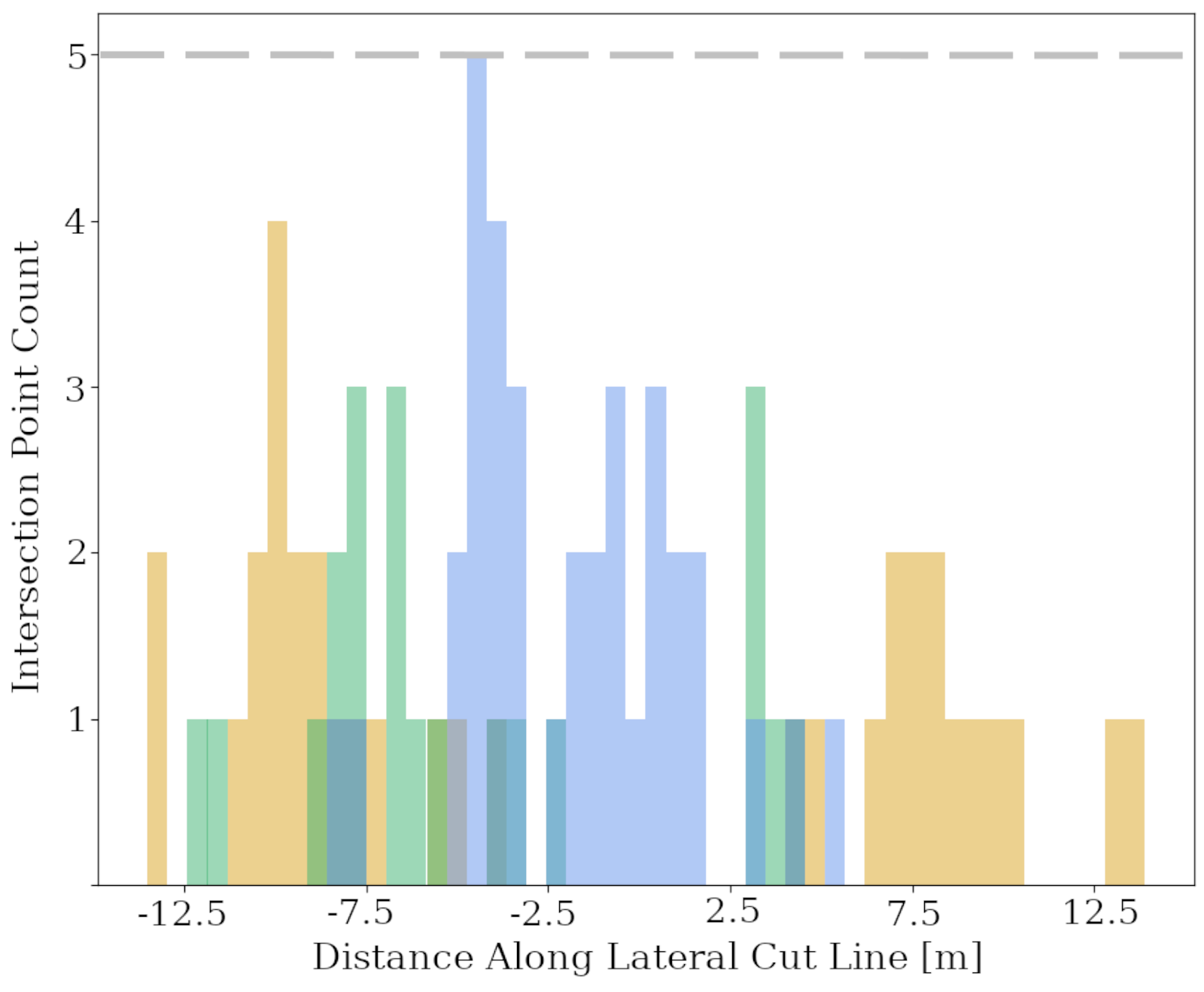}
        \caption{Before lateral correction}
        \label{fig:histograms_lateral_optimization_before}
    \end{subfigure}
    \begin{subfigure}[c]{\columnwidth}
        \centering
        \includegraphics[width=\columnwidth]{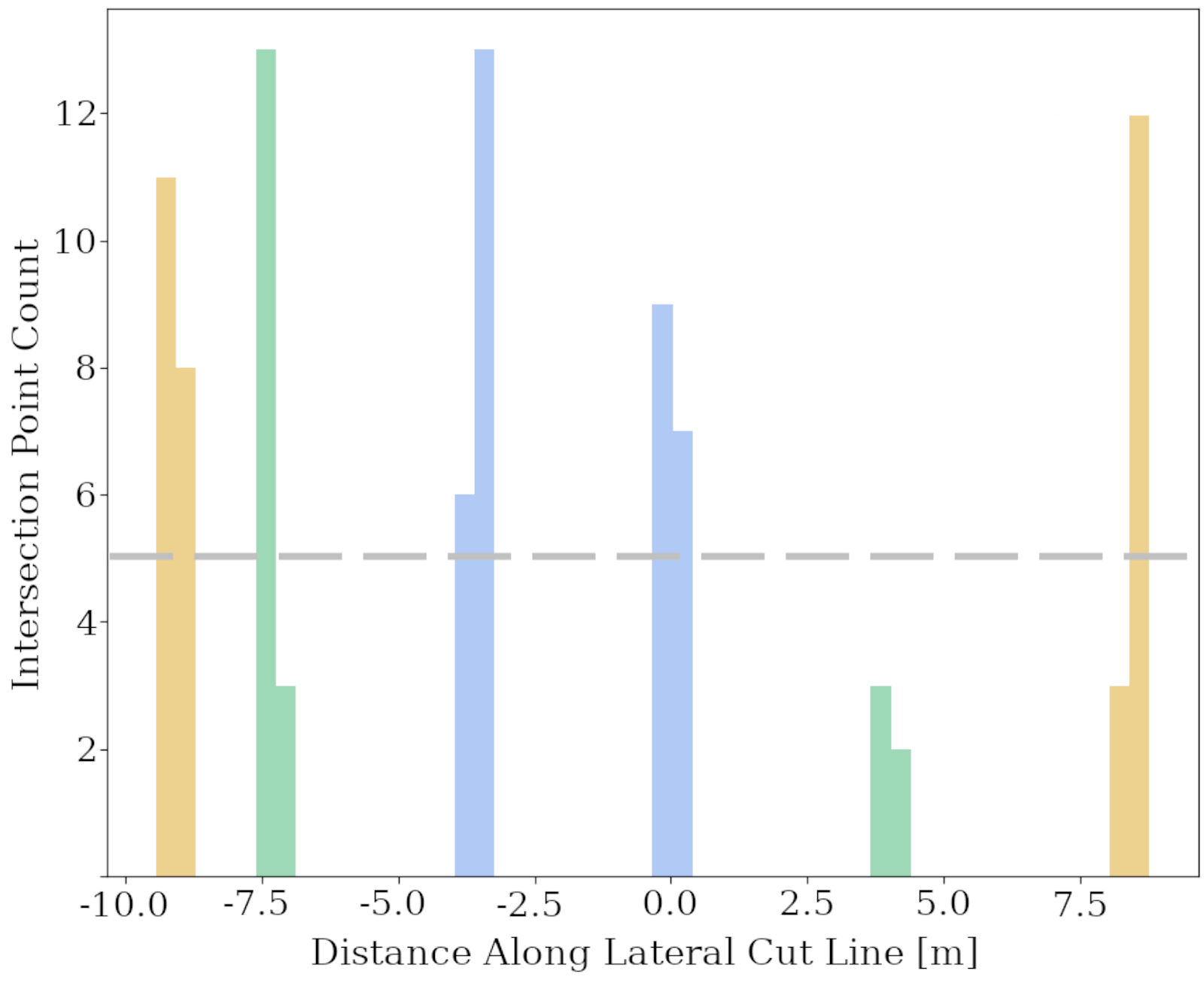}
        \caption{After lateral correction}
        \label{fig:histograms_lateral_optimization_after}
    \end{subfigure}
    \caption{Histogram of sampled detection points before and after EM-based lateral correction.
        Road boundary samples are shown in yellow, dashed lane marking samples in blue, solid lane marking samples in green.}
    \label{fig:histograms_lateral_optimization}
\end{figure*}

Fig. \ref{fig:histograms_lateral_optimization_before} shows a distribution of sampled detection points before the lateral correction.
The large spread of detections needs to be corrected for sensible aggregation of the measurements.
We therefore try to minimize the alignment error between the different drives, assuming drive localization errors as the main source of misalignment rather than individual measurement noise. Referring back to this assumption, we estimate one lateral offset for each drive as correction for the localization.
This means that all detection points belonging to a single drive can only be moved in the same direction and with the same offset.

As the co-referencing of detections, meaning the assignment of detections to the same road feature, is interdependent with the lateral offset estimation, we use an iterative EM approach.
First, the detections are separated by type and clustered using the k-means algorithm, using the silhouette score as the selection criterion for the number of clusters.
We note

\begin{align}
    \mathbf{o} = \left(o_1, o_2, \dots, o_m\right)
\end{align}

as the vector of $m$ drive offsets, with one offset assigned per drive and all offsets parametrized using the cut line vector. The lateral offset for each drive is then estimated by minimizing the cost function given below:

\begin{align}
    \label{eq:cost_function}
    \mathbf{o_{opt}}  & = \argmin_{\mathbf{o}} \sum_{i=1}^n \text{Var}\left(c_i\left(\mathbf{o}\right)\right) + \sum_{i=1}^n f\left(c_i\left(\mathbf{o}\right)\right) \\
    f\left(c_i\right) & = e^{-s \cdot {d_{min}}^2}
\end{align}

Where $\text{Var}\left(c_i\left(\mathbf{o}\right)\right)$ denotes the variance of cluster $c_i$ corrected with $\mathbf{o}$. $f\left(c_i\right)$ describes a penalty function that increases for small distances to the nearest cluster of the same type $d_{min}$ and can be adjusted depending on the expected lane width with the parameter $s$.
The described optimization problem can be solved with any non-linear least squares algorithm, e.g. Levenberg-Marquardt.
Finally, the detections are re-clustered using the new offsets and the process is repeated for a fixed number of iterations.

Fig. \ref{fig:histograms_lateral_optimization_after} displays the result of the EM optimization on the points in Fig. \ref{fig:histograms_lateral_optimization_before} and shows how overlapping detection clusters with large spread can be aligned and separated, also exemplifying localization errors as the main source of misalignment.
Optimization results are scored with their silhouette score, and results below a threshold of 0.67 are rejected and the respective sampling step is skipped to prevent wrongly aggregated points from being used in later steps of the pipeline. Additionally, possible drift of the optimization results compared to the actual road shape is corrected by aligning the road midpoint post-optimization with the one pre-optimization.

\subsection{Aggregation of Clustered Detection Points}
\label{sec:aggregation_of_clustered_detection_points}

After sec. \ref{sec:expectation_maximization}, we have a set of clustered and aligned detection points.
To aggregate the measurements of a single cluster, we estimate its PDF using Kernel Density Estimation.
The KDE is estimated using a Gaussian kernel with a bandwidth of 6 times the clusters standard deviation.
The detection points belonging to the same cluster are then replaced by the maximum of the estimated PDF.
If there are misclassifications of road feature detections present, estimated PDFs of different clusters may overlap each other, which can be used to filter out these misclassifications. If two PDFs overlap area exceeds a certain threshold, the cluster with the lower number of detection points is removed.

The calculated maximum points of the PDFs now each represent the aggregated detections of a single road feature.
They are saved for every drive whose detections were included in the optimization, with the traveled distance along the trajectory of the respective drive as the key.
With this information, points on drives that were already processed during another pivot drive do not need to be calculated again, saving large amounts of computation time.
The whole process described in this section is repeated for every drive in the input data.

\section{Connection of Distribution Maxima}
\label{sec:connection_of_distribution_maxima}

The difficult problem of connecting arbitrary point sets to form semantically useful polylines can be massively simplified with the contextual information of the lateral sampling along the associated pivot drive and thus the ego-road.
Since we know that the order of point sets follows the trajectory of the drive, we can infer that the aggregated points of succeeding point sets must also be connected.
The hard problem of finding polylines can therefore be separated into finding the optimal connections between succeeding point sets, which is a rectangular linear assignment problem for bipartite graphs.
This problem can be solved efficiently using the algorithm described in \cite{Karp1980}, using the distance between the points of opposite sets as the edge weights.

However, the algorithm does not allow for the assignment of leftover nodes, which can occur at lane splits and merges, where the number of points is not equal on both sides of the bipartite graph.
To solve this problem, we use a modified version of the algorithm, where the leftover nodes are assigned to their closest neighbor on the other side of the bipartite graph.
This approximates the topological connections commonly encountered for road markings.

\begin{figure}[htb]
    \includegraphics[width=\columnwidth]{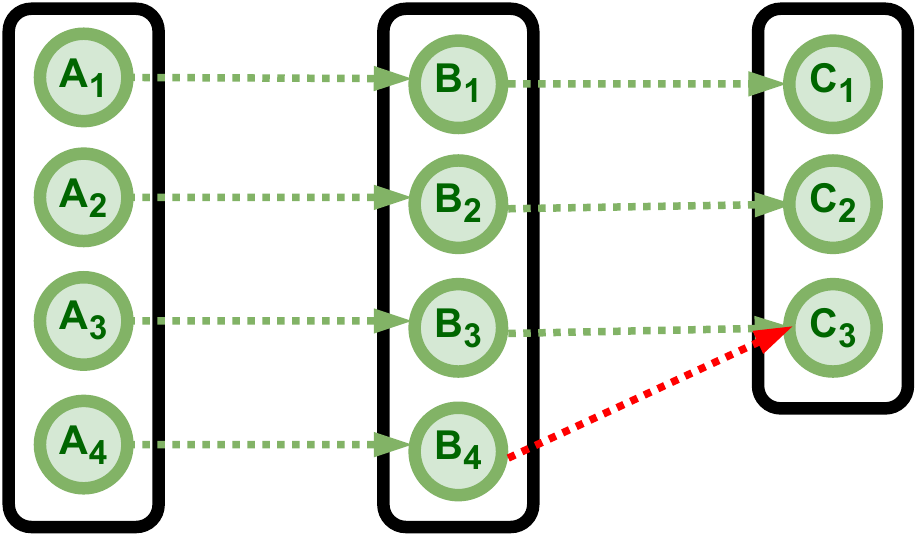}
    \caption{Example of assignments of the used modified rectangular assignment algorithm \cite{Karp1980}. The leftover node $B_4$ is assigned to its closest bipartite neighbor $C_3$.}
    \label{fig:hungarian_example}
    \vspace{-2mm}
\end{figure}

Figure \ref{fig:hungarian_example} shows an example of the modified algorithm, where the unmodified algorithm produces the assignment shown by the green arrows, with node $B_4$ left unassigned. The modified algorithm assigns leftover nodes to their closest neighbor on the other side of the bipartite graph, assigning node $B_4$ to node $C_3$, as shown by the red arrow.

This procedure is performed sequentially for each point set, filtering out implausible optimization results via an average connection angle threshold per bipartite graph as well as preventing connections of detections of a different type to road borders.
As described in Sec. \ref{sec:expectation_maximization}, failure of the optimization may lead to patches with no aggregated detections, which can be bridged by connecting to the next valid point set, if the distance is below a given threshold.

To receive polylines out of the calculated graph edges, a set of open and closed polylines is maintained.
Polylines are opened for the starting points of a drive.
They are closed and a new polyline is opened in the following conditions:

\begin{itemize}
    \item The connection edge connects two detection points of a different type, e.g. a dashed line and a solid line.
    \item The connection is made after the regular rectangular assignment algorithm (leftover node, as in $B_4$ to node $C_3$ in Fig. \ref{fig:hungarian_example})
    \item The distance between the two detection point sets exceeds a given threshold. This can happen when the optimization fails to find a solution for a long road segment.
    \item We reach a section that was processed by a different pivot drive.
\end{itemize}
\section {Inter-Segment Connection}
\label{sec:inter_segment_connection}

\begin{figure}[htb]
    \includegraphics[width=\columnwidth]{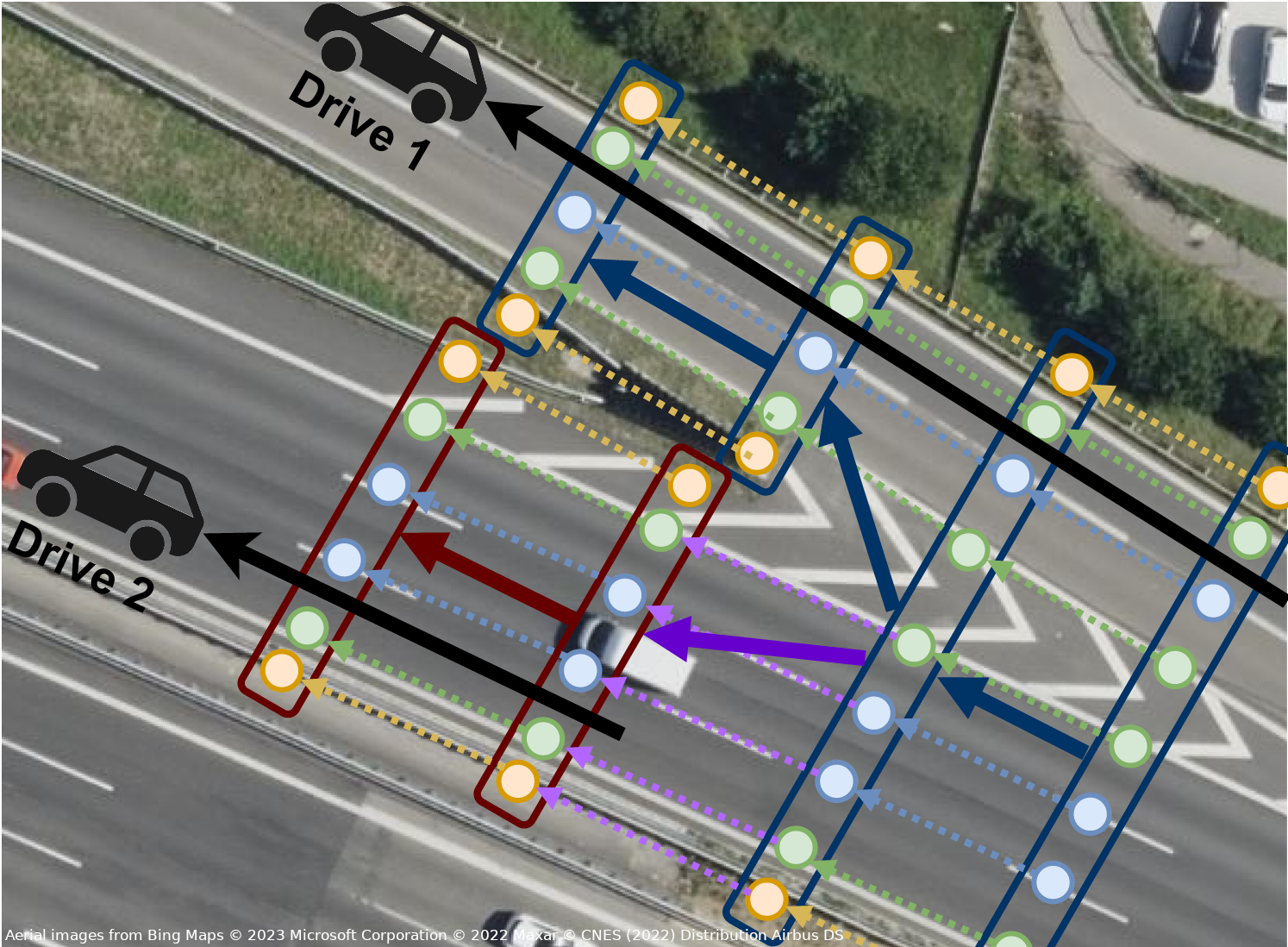}
    \caption{Visualization of the hierarchical graph structure used for stitching of map segments. The high-level topology graph (blue/red) contains information about the road topology, while the low-level road feature graph (colored nodes) contains the actual road feature polylines. Nodes in both graphs contain references to each other. Two segments (blue and red) from different drives can be stitched together as shown with the purple edges.}
    \label{fig:hierarchical_graphs}
    \vspace{-2mm}
\end{figure}

Sec. \ref{sec:connection_of_distribution_maxima} results in individually connected segments for every drive.
These segments are stitched together into a complete map using a hierarchical graph structure comprised of a high-level topology graph and a low-level road feature graph.
This structure is created for every segment during Sec. \ref{sec:connection_of_distribution_maxima} as well as for the final complete map graph.

Fig. \ref{fig:hierarchical_graphs} shows a visualization of this structure.
A node in the topology graph, shown by the red and blue rectangles, is created for each sampled step in Sec. \ref{sec:lateral_sampling_of_detection_points} and contains the road midpoint at its position as well as a reference to the aggregated detection points of its step.
The topology graph can be seen as similar to a regular SD map, as it contains information about the high-level road topology.
The low-level road feature graph (the small circles in Fig. \ref{fig:hierarchical_graphs}) contains the actual road feature polylines, which are created by connecting the aggregated detection points of the topology graph nodes.
To insert the results of a new drive into the complete graph, visualized by the purple arrows, the following steps are performed:

\begin{enumerate}
    \item The drive topology graph is separated into its connected components. Breaks in the connection may for example occur in sections that have already been processed by another pivot drive.
    \item The start and end nodes of a connected component are extracted and their nearest neighbor nodes in the complete topology graph are found using a kd-tree.
    \item The possible connection to the neighbor node is validated with a distance and connection angle check.
    \item If successful, the connection to the neighbor node is added to the complete topology graph and the low-level road feature polylines are also connected using the referenced low-level nodes in the topology nodes. The method for connecting the polylines is as described in Sec. \ref{sec:connection_of_distribution_maxima}, with the addition that only nodes of the same road feature type are connected and that the unmodified version of the rectangular assignment algorithm in \cite{Karp1980} is used.
\end{enumerate}

The system proposed here allows for the creation of a map with entire highway networks, including exit and entry ramps, as well as updating an existing map with newly mapped sections. %

\section{Evaluation}
\label{sec:Evaluation}

To evaluate the performance of the proposed system, we use a dataset of vehicle fleet data from 134 drive snippets on a highway section near Stuttgart, Germany.
The section consists of two opposite highway directions with a total length of 29.1 km, including highway exit and entry ramps.

\subsection{Ground Truth Dataset}
\label{sec:GroundTruthAnnotation}

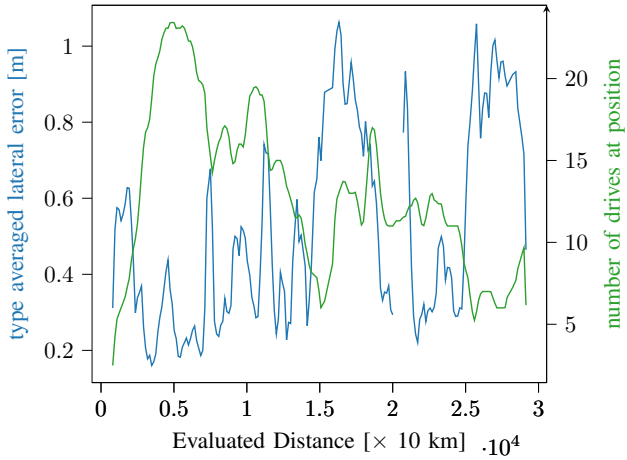
\begin{figure}
    \centering

    \resizebox{\columnwidth}{!}{
        \begin{tikzpicture}

            \definecolor{darkgray176}{RGB}{176,176,176}
            \definecolor{forestgreen4416044}{RGB}{44,160,44}
            \definecolor{lightgray204}{RGB}{204,204,204}
            \definecolor{steelblue31119180}{RGB}{31,119,180}

            \begin{axis}[
                    legend style={fill opacity=0.8, draw opacity=1, text opacity=1, draw=lightgray204},
                    tick align=outside,
                    tick pos=left,
                    unbounded coords=jump,
                    x grid style={darkgray176},
                    xlabel={Evaluated Distance [$\times$ 10 km]},
                    xmin=-614.8, xmax=30554.8,
                    xtick style={color=black},
                    y grid style={darkgray176},
                    ylabel=\textcolor{steelblue31119180}{type averaged lateral error [m]},
                    ymin=0.115255706812259, ymax=1.10744090842631,
                    ytick style={color=black}
                ]
                \addplot [semithick, steelblue31119180, forget plot]
                table [row sep=crcr] {%
                        802 0.311526467465721\\
                        956 0.517514233730289\\
                        1096 0.575208109124237\\
                        1236 0.5701875555063\\
                        1376 0.541358165856481\\
                        1516 0.558202611903157\\
                        1656 0.585092120528803\\
                        1796 0.627824375872315\\
                        1936 0.626090072749385\\
                        2076 0.542745594100215\\
                        2216 0.435905302831008\\
                        2356 0.298667892522708\\
                        2496 0.338280150100577\\
                        2636 0.352089557497\\
                        2776 0.368948039983448\\
                        2916 0.26192492916543\\
                        3056 0.206179062626544\\
                        3196 0.177712951246821\\
                        3336 0.187055729275211\\
                        3476 0.160355034158352\\
                        3616 0.170360060727436\\
                        3756 0.190872069376445\\
                        3896 0.247807218167994\\
                        4036 0.287318556357063\\
                        4176 0.315322865740013\\
                        4316 0.362034856969829\\
                        4456 0.407501578835567\\
                        4596 0.437681271624553\\
                        4736 0.359485345918646\\
                        4876 0.323925706589946\\
                        5016 0.252316788350771\\
                        5156 0.231061067741765\\
                        5296 0.184286272835431\\
                        5436 0.182028300784678\\
                        5576 0.20839435810384\\
                        5716 0.220243994153172\\
                        5856 0.23249334211714\\
                        5996 0.214315415290849\\
                        6136 0.230817646662012\\
                        6276 0.253152000081413\\
                        6418 0.263993833742835\\
                        6558 0.22555804875853\\
                        6698 0.205778224774629\\
                        6838 0.186675610957932\\
                        6978 0.199755282346775\\
                        7118 0.309342613868817\\
                        7258 0.599548750357786\\
                        7498 0.677511940649891\\
                        7638 0.522267089649455\\
                        7778 0.276885868101511\\
                        7930 0.242848125954531\\
                        8070 0.236788914972372\\
                        8210 0.266441824706714\\
                        8350 0.275930938912094\\
                        8490 0.336428331278865\\
                        8630 0.302112301852665\\
                        8770 0.297637508474858\\
                        8910 0.318080379405822\\
                        9050 0.466140159352629\\
                        9190 0.49996897895861\\
                        9330 0.493994066660685\\
                        9470 0.449211467507903\\
                        9610 0.524964475647695\\
                        9750 0.515718635218646\\
                        9890 0.495467263109726\\
                        10030 0.434882407962456\\
                        10202 0.397334506199196\\
                        10342 0.339617082935093\\
                        10482 0.317330114863589\\
                        10622 0.285678062053218\\
                        10762 0.288963369476264\\
                        10902 0.395151185870136\\
                        11042 0.533495073021231\\
                        11182 0.741134526933965\\
                        11322 0.71914120950999\\
                        11462 0.717458653215373\\
                        11602 0.578340692937976\\
                        11742 0.425057036955125\\
                        11882 0.303969531262052\\
                        12022 0.24296374215714\\
                        12162 0.283392348420771\\
                        12302 0.404912889747497\\
                        12442 0.377652977765387\\
                        12582 0.35525319207173\\
                        12722 0.22760905717682\\
                        12862 0.273649159937319\\
                        13002 0.270732248778006\\
                        13142 0.431455922126695\\
                        13300 0.526232881025885\\
                        13440 0.596920933703387\\
                        13580 0.48671556453331\\
                        13720 0.502387653069604\\
                        13860 0.462525821714862\\
                        14000 0.421656761762328\\
                        14140 0.264782925395693\\
                        14280 0.333166577918166\\
                        14420 0.419826499399851\\
                        14622 0.615765775397489\\
                        14800 0.650752284889774\\
                        14940 0.760877177908047\\
                        15080 0.698966397890318\\
                        15330 0.878781373550714\\
                        15874 0.890745575815126\\
                        16036 0.992452095543537\\
                        16176 1.03925979566241\\
                        16316 1.06234158108022\\
                        16456 1.02903587536065\\
                        16596 0.902106156209835\\
                        16736 0.847219412332667\\
                        16876 0.84887302782235\\
                        17016 0.902773228974178\\
                        17156 0.957009478804262\\
                        17296 0.913602689327223\\
                        17436 0.858022622696252\\
                        17576 0.832535206600621\\
                        17716 0.78769939259356\\
                        17856 0.770160656623868\\
                        17996 0.710443549922311\\
                        18136 0.801633302420229\\
                        18314 0.725883747947028\\
                        18464 0.73835781142503\\
                        18604 0.63252887424414\\
                        18744 0.644638678211689\\
                        18884 0.587670780020456\\
                        19024 0.489313214400593\\
                        19164 0.363093544808026\\
                        19304 0.331611939780085\\
                        19444 0.353378101533562\\
                        19584 0.350325182629714\\
                        19724 0.369653754172451\\
                        19864 0.306662144022073\\
                        20004 0.293458271472275\\
                        20244 nan\\
                        20492 nan\\
                        20732 0.772730125560617\\
                        20872 0.933441782529492\\
                        21012 0.837425918122561\\
                        21152 0.616144030661555\\
                        21292 0.413501576652596\\
                        21432 0.289791430241283\\
                        21572 0.242362402766862\\
                        21712 0.220726197587756\\
                        21852 0.281797972819661\\
                        21992 0.29451695451685\\
                        22132 0.321174839958919\\
                        22272 0.275919683238133\\
                        22412 0.304290265211192\\
                        22564 0.31141399525581\\
                        22704 0.301670561822312\\
                        22844 0.31297263102656\\
                        22984 0.349865918103092\\
                        23124 0.469199180333703\\
                        23338 0.499007155322939\\
                        23478 0.473626830287841\\
                        23618 0.38495273952694\\
                        23758 0.418126412123043\\
                        23898 0.416688393529955\\
                        24038 0.391636651478808\\
                        24178 0.303718142116361\\
                        24318 0.28936924832084\\
                        24458 0.290324743626296\\
                        24598 0.311804035430385\\
                        24738 0.308936551536529\\
                        24878 0.426162490333156\\
                        25018 0.592270521007626\\
                        25158 0.737703358199782\\
                        25298 0.797994919694375\\
                        25466 0.883995256294509\\
                        25606 0.978173580176183\\
                        25746 1.05863339655418\\
                        25886 0.904796650035681\\
                        26026 0.757971751750826\\
                        26166 0.836008274226268\\
                        26306 0.876814955232292\\
                        26446 0.875605625281653\\
                        26586 0.812259437287106\\
                        26726 0.930574242299055\\
                        26866 1.00080155814258\\
                        27006 1.01613828906472\\
                        27146 0.962378999090287\\
                        27286 0.916491855936551\\
                        27426 0.958555000389347\\
                        27566 0.960392595775443\\
                        27706 0.921234075771889\\
                        27846 0.895429173820786\\
                        28226 0.925159407388283\\
                        28440 0.932534883567107\\
                        28580 0.835753858053745\\
                        28858 0.761521199059709\\
                        28998 0.715514884945712\\
                        29138 0.464342387200613\\
                    };
            \end{axis}

            \begin{axis}[
                    axis y line=right,
                    tick align=outside,
                    x grid style={darkgray176},
                    xmin=-614.8, xmax=30554.8,
                    xtick pos=left,
                    xtick style={color=black},
                    y grid style={darkgray176},
                    ylabel=\textcolor{forestgreen4416044}{number of drives at position},
                    ymin=1.43775, ymax=24.47725,
                    ytick pos=right,
                    ytick style={color=black},
                    yticklabel style={anchor=west}
                ]
                \addplot [semithick, forestgreen4416044]
                table [row sep=crcr]{%
                        802 2.485\\
                        956 4.235\\
                        1096 5.315\\
                        1236 5.665\\
                        1376 6\\
                        1516 6.325\\
                        1656 6.675\\
                        1796 7.25\\
                        1936 7.805\\
                        2076 8.85500000000001\\
                        2216 9.605\\
                        2356 10.23\\
                        2496 11.11\\
                        2636 12.53\\
                        2776 14.28\\
                        2916 15.8\\
                        3056 16.73\\
                        3196 17\\
                        3336 17.515\\
                        3476 18.755\\
                        3616 20.15\\
                        3756 20.89\\
                        3896 21.265\\
                        4036 21.705\\
                        4176 22.405\\
                        4316 22.79\\
                        4456 23.075\\
                        4596 23.075\\
                        4736 23.4\\
                        4876 23.43\\
                        5016 23.405\\
                        5156 23.055\\
                        5296 23.085\\
                        5436 23.085\\
                        5576 23.215\\
                        5716 23.08\\
                        5856 22.73\\
                        5996 22.295\\
                        6136 22.285\\
                        6276 21.935\\
                        6418 21.28\\
                        6558 20.355\\
                        6698 19.88\\
                        6838 19.79\\
                        6978 19.615\\
                        7118 19.105\\
                        7258 17.015\\
                        7498 15.255\\
                        7638 14.26\\
                        7778 14.99\\
                        7930 15.64\\
                        8070 16.115\\
                        8210 16.395\\
                        8350 16.93\\
                        8490 17.115\\
                        8630 16.95\\
                        8770 16.445\\
                        8910 15.395\\
                        9050 14.81\\
                        9190 14.93\\
                        9330 15.63\\
                        9470 16\\
                        9610 16\\
                        9750 15.825\\
                        9890 16.535\\
                        10030 17.73\\
                        10202 18.95\\
                        10342 19.15\\
                        10482 19.445\\
                        10622 19.5\\
                        10762 19.355\\
                        10902 19.005\\
                        11042 19\\
                        11182 18.6\\
                        11322 17.225\\
                        11462 15.475\\
                        11602 14.395\\
                        11742 14.46\\
                        11882 14.81\\
                        12022 15\\
                        12162 15\\
                        12302 15\\
                        12442 14.69\\
                        12582 14.125\\
                        12722 13.435\\
                        12862 13.07\\
                        13002 12.635\\
                        13142 12.185\\
                        13300 11.55\\
                        13440 11.495\\
                        13580 11.675\\
                        13720 11.49\\
                        13860 10.795\\
                        14000 9.875\\
                        14140 9.27999999999999\\
                        14280 8.57999999999999\\
                        14420 8.17\\
                        14622 8\\
                        14800 7.84\\
                        14940 7.07\\
                        15080 6.02\\
                        15330 6.41\\
                        15874 8.68\\
                        16036 11.405\\
                        16176 12.66\\
                        16316 13.03\\
                        16456 13.405\\
                        16596 13.72\\
                        16736 13.695\\
                        16876 13.345\\
                        17016 13\\
                        17156 13\\
                        17296 13\\
                        17436 13.055\\
                        17576 12.52\\
                        17716 11.82\\
                        17856 11.065\\
                        17996 11.27\\
                        18136 12.725\\
                        18314 14.755\\
                        18464 16.555\\
                        18604 17\\
                        18744 16.865\\
                        18884 15.925\\
                        19024 14.51\\
                        19164 12.99\\
                        19304 12.06\\
                        19444 11.425\\
                        19584 11.075\\
                        19724 11\\
                        19864 11\\
                        20004 11\\
                        20244 11.315\\
                        20492 11.315\\
                        20732 11.54\\
                        20872 11.585\\
                        21012 11.935\\
                        21152 12.23\\
                        21292 12.23\\
                        21432 12.57\\
                        21572 12.695\\
                        21712 12.425\\
                        21852 11.725\\
                        21992 11.095\\
                        22132 11.02\\
                        22272 11.47\\
                        22412 12.15\\
                        22564 12.845\\
                        22704 12.975\\
                        22844 12.59\\
                        22984 12.495\\
                        23124 12.365\\
                        23338 12.335\\
                        23478 11.68\\
                        23618 11.235\\
                        23758 11\\
                        23898 11\\
                        24038 11\\
                        24178 11\\
                        24318 11\\
                        24458 11\\
                        24598 10.54\\
                        24738 9.80999999999999\\
                        24878 8.79499999999999\\
                        25018 8.26999999999999\\
                        25158 7.835\\
                        25298 6.785\\
                        25466 5.735\\
                        25606 5.24\\
                        25746 5.75\\
                        25886 6.45\\
                        26026 6.86\\
                        26166 7\\
                        26306 7\\
                        26446 7\\
                        26586 7\\
                        26726 6.985\\
                        26866 6.635\\
                        27006 6.285\\
                        27146 6\\
                        27286 6\\
                        27426 6\\
                        27566 6\\
                        27706 6.015\\
                        27846 6.635\\
                        28226 7.245\\
                        28440 7.665\\
                        28580 8.395\\
                        28858 9.26\\
                        28998 9.67000000000001\\
                        29138 6.17\\
                    };
            \end{axis}

        \end{tikzpicture}
    }
    \caption{Average lateral error over driving distance in relation to the number of
        drives traversing that location in the dataset. This data is sliding window low-pass filtered for clear visibility. Number of drives and total error correlate inversely. }
    \label{fig:error_over_meters}
    \vspace{-2mm}
\end{figure}

We evaluate the lateral error of the resulting map by comparing it to a hand-labeled ground truth dataset generated from satellite imagery.
The ground-truth data is validated with multiple imagery sources, eliminating parts of the highway for evaluation where recent construction sites hinder accurate ground-truth annotation.
The total length of annotated polylines is 204 km.
The fleet data has a bias towards one direction, showing a higher number of drives in this direction as visible in the first 15 kilometers of fig. \ref{fig:error_over_meters}.

The generated ground truth additionally provides a reference line and region-of-interest in comparison to the sampling of pivot drive trajectories and cut line refinement during the optimization step.
Like in section \ref{sec:stepwise_expectation_maximization}, a lateral cut line is computed for every two meters along the reference line.
This cut line is restricted by the region-of-interest borders and all intersecting polylines in the resulting map and ground truth data are retrieved.
We associate an intersection point of the mapping result with the closest ground truth point and compute the distance between measurement and ground truth for each map feature type.
The influencing factors on the resulting lateral distance error are investigated in the remainder of this section.

\subsection{Evaluation Metrics}
\label{sec:EvaluationMetrics}

\begin{figure}[htb]
\def\svgwidth{\columnwidth}
\begingroup%
\makeatletter%
\providecommand\color[2][]{%
    \errmessage{(Inkscape) Color is used for the text in Inkscape, but the package 'color.sty' is not loaded}%
    \renewcommand\color[2][]{}%
}%
\providecommand\transparent[1]{%
    \errmessage{(Inkscape) Transparency is used (non-zero) for the text in Inkscape, but the package 'transparent.sty' is not loaded}%
    \renewcommand\transparent[1]{}%
}%
\providecommand\rotatebox[2]{#2}%
\newcommand*\fsize{\dimexpr\f@size pt\relax}%
\newcommand*\lineheight[1]{\fontsize{\fsize}{#1\fsize}\selectfont}%
\ifx\svgwidth\undefined%
    \setlength{\unitlength}{836.16000366bp}%
    \ifx\svgscale\undefined%
        \relax%
    \else%
        \setlength{\unitlength}{\unitlength * \real{\svgscale}}%
    \fi%
\else%
    \setlength{\unitlength}{\svgwidth}%
\fi%
\global\let\svgwidth\undefined%
\global\let\svgscale\undefined%
\makeatother%
\begin{picture}(1,0.56429391)%
    \lineheight{1}%
    \setlength\tabcolsep{0pt}%
    \put(0,0){\includegraphics[width=\unitlength,page=1]{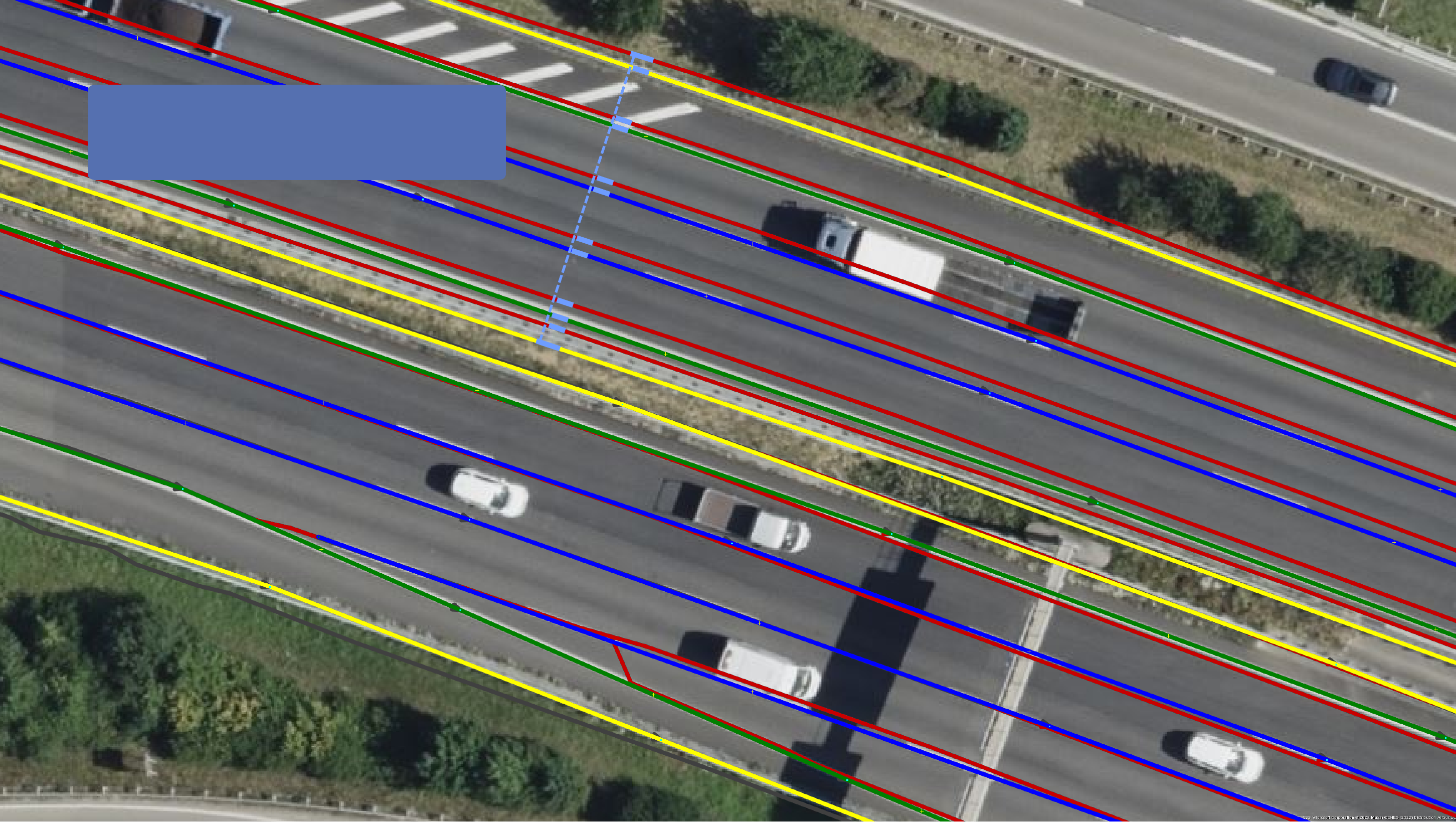}}%
    \put(0.0702666,0.45803467){\color[rgb]{1,1,1}\makebox(0,0)[lt]{\lineheight{1.25}\smash{\begin{tabular}[t]{l}$\Delta$ offset error\end{tabular}}}}%
    \put(0,0){\includegraphics[width=\unitlength,page=2]{GT_and_Mapped_Result.pdf}}%
    \put(0.15758047,0.15155528){\makebox(0,0)[lt]{\lineheight{1.25}\smash{\begin{tabular}[t]{l}$\Delta$ non-offset error\end{tabular}}}}%
    \put(0,0){\includegraphics[width=\unitlength,page=3]{GT_and_Mapped_Result.pdf}}%
\end{picture}%
\endgroup%
\caption{Visualization of the evaluation metrics. The ground truth with road boundaries is shown in yellow, dashed lines in blue and solid lines in green. The mapped result is shown in red, without type color coding. The offset error, resulting from localization errors, is indicated by the blue arrow. As these errors are related to a small number of drives in some sections of the evaluated dataset, we also evaluate a offset corrected error. A sample of remaining uncorrectable non-offset error is visualized in orange.}
\label{fig:evaluation_metrics}
\vspace{-2mm}
\end{figure}

\begin{figure}
    \centering
    \begin{tikzpicture}

        \definecolor{darkgray176}{RGB}{176,176,176}
        \definecolor{darkslategray63}{RGB}{63,63,63}
        \definecolor{lightgray204}{RGB}{204,204,204}
        \definecolor{peru22412844}{RGB}{224,128,44}
        \definecolor{steelblue49115161}{RGB}{49,115,161}

        \begin{axis}[
                legend cell align={left},
                legend style={
                        fill opacity=0.8,
                        draw opacity=1,
                        text opacity=1,
                        at={(0.03,0.97)},
                        anchor=north west,
                        draw=lightgray204
                    },
                tick align=outside,
                tick pos=left,
                x grid style={darkgray176},
                xmin=-0.5, xmax=2.5,
                xtick style={color=black},
                xtick={0,1,2},
                xticklabels={Road Border,Solid Line,Dashed Line},
                y grid style={darkgray176},
                ylabel={lateral error [m]},
                ymin=-0.0824677554097366, ymax=1.73182286360447,
                ytick style={color=black}
            ]
            \path [draw=darkslategray63, fill=steelblue49115161, semithick]
            (axis cs:-0.396,0.261840578319356)
            --(axis cs:-0.004,0.261840578319356)
            --(axis cs:-0.004,0.743366885881045)
            --(axis cs:-0.396,0.743366885881045)
            --(axis cs:-0.396,0.261840578319356)
            --cycle;
            \path [draw=darkslategray63, fill=peru22412844, semithick]
            (axis cs:0.004,0.168225089187773)
            --(axis cs:0.396,0.168225089187773)
            --(axis cs:0.396,0.433723526037922)
            --(axis cs:0.004,0.433723526037922)
            --(axis cs:0.004,0.168225089187773)
            --cycle;
            \path [draw=darkslategray63, fill=steelblue49115161, semithick]
            (axis cs:0.604,0.211150549975115)
            --(axis cs:0.996,0.211150549975115)
            --(axis cs:0.996,0.667891995642148)
            --(axis cs:0.604,0.667891995642148)
            --(axis cs:0.604,0.211150549975115)
            --cycle;
            \path [draw=darkslategray63, fill=peru22412844, semithick]
            (axis cs:1.004,0.13002752020773)
            --(axis cs:1.396,0.13002752020773)
            --(axis cs:1.396,0.244575346285976)
            --(axis cs:1.004,0.244575346285976)
            --(axis cs:1.004,0.13002752020773)
            --cycle;
            \path [draw=darkslategray63, fill=steelblue49115161, semithick]
            (axis cs:1.604,0.163071167770206)
            --(axis cs:1.996,0.163071167770206)
            --(axis cs:1.996,0.758610247150457)
            --(axis cs:1.604,0.758610247150457)
            --(axis cs:1.604,0.163071167770206)
            --cycle;
            \path [draw=darkslategray63, fill=peru22412844, semithick]
            (axis cs:2.004,0.053040672513861)
            --(axis cs:2.396,0.053040672513861)
            --(axis cs:2.396,0.149338805230651)
            --(axis cs:2.004,0.149338805230651)
            --(axis cs:2.004,0.053040672513861)
            --cycle;
            \draw[draw=darkslategray63,fill=steelblue49115161,line width=0.3pt] (axis cs:0,0) rectangle (axis cs:0,0);
            \addlegendimage{ybar,ybar legend,draw=darkslategray63,fill=steelblue49115161,line width=0.3pt}

            \draw[draw=darkslategray63,fill=peru22412844,line width=0.3pt] (axis cs:0,0) rectangle (axis cs:0,0);
            \addlegendimage{ybar,ybar legend,draw=darkslategray63,fill=peru22412844,line width=0.3pt}

            \addplot [semithick, darkslategray63, forget plot]
            table {%
                    -0.2 0.261840578319356
                    -0.2 0.000151856014134849
                };
            \addplot [semithick, darkslategray63, forget plot]
            table {%
                    -0.2 0.743366885881045
                    -0.2 1.46452568046912
                };
            \addplot [semithick, darkslategray63, forget plot]
            table {%
                    -0.298 0.000151856014134849
                    -0.102 0.000151856014134849
                };
            \addplot [semithick, darkslategray63, forget plot]
            table {%
                    -0.298 1.46452568046912
                    -0.102 1.46452568046912
                };
            \addplot [semithick, darkslategray63, forget plot]
            table {%
                    0.2 0.168225089187773
                    0.2 0
                };
            \addplot [semithick, darkslategray63, forget plot]
            table {%
                    0.2 0.433723526037922
                    0.2 0.83168759141118
                };
            \addplot [semithick, darkslategray63, forget plot]
            table {%
                    0.102 0
                    0.298 0
                };
            \addplot [semithick, darkslategray63, forget plot]
            table {%
                    0.102 0.83168759141118
                    0.298 0.83168759141118
                };
            \addplot [semithick, darkslategray63, forget plot]
            table {%
                    0.8 0.211150549975115
                    0.8 0.000815781208947452
                };
            \addplot [semithick, darkslategray63, forget plot]
            table {%
                    0.8 0.667891995642148
                    0.8 1.35289147534926
                };
            \addplot [semithick, darkslategray63, forget plot]
            table {%
                    0.702 0.000815781208947452
                    0.898 0.000815781208947452
                };
            \addplot [semithick, darkslategray63, forget plot]
            table {%
                    0.702 1.35289147534926
                    0.898 1.35289147534926
                };
            \addplot [semithick, darkslategray63, forget plot]
            table {%
                    1.2 0.13002752020773
                    1.2 0
                };
            \addplot [semithick, darkslategray63, forget plot]
            table {%
                    1.2 0.244575346285976
                    1.2 0.416325645174781
                };
            \addplot [semithick, darkslategray63, forget plot]
            table {%
                    1.102 0
                    1.298 0
                };
            \addplot [semithick, darkslategray63, forget plot]
            table {%
                    1.102 0.416325645174781
                    1.298 0.416325645174781
                };
            \addplot [semithick, darkslategray63, forget plot]
            table {%
                    1.8 0.163071167770206
                    1.8 0.000856396362733094
                };
            \addplot [semithick, darkslategray63, forget plot]
            table {%
                    1.8 0.758610247150457
                    1.8 1.64935510819473
                };
            \addplot [semithick, darkslategray63, forget plot]
            table {%
                    1.702 0.000856396362733094
                    1.898 0.000856396362733094
                };
            \addplot [semithick, darkslategray63, forget plot]
            table {%
                    1.702 1.64935510819473
                    1.898 1.64935510819473
                };
            \addplot [semithick, darkslategray63, forget plot]
            table {%
                    2.2 0.053040672513861
                    2.2 5.82076609134674e-11
                };
            \addplot [semithick, darkslategray63, forget plot]
            table {%
                    2.2 0.149338805230651
                    2.2 0.29351866939432
                };
            \addplot [semithick, darkslategray63, forget plot]
            table {%
                    2.102 5.82076609134674e-11
                    2.298 5.82076609134674e-11
                };
            \addplot [semithick, darkslategray63, forget plot]
            table {%
                    2.102 0.29351866939432
                    2.298 0.29351866939432
                };
            \addplot [semithick, darkslategray63, forget plot]
            table {%
                    -0.396 0.433061009620325
                    -0.004 0.433061009620325
                };
            \addplot [semithick, darkslategray63, forget plot]
            table {%
                    0.004 0.266639106332247
                    0.396 0.266639106332247
                };
            \addplot [semithick, darkslategray63, forget plot]
            table {%
                    0.604 0.343853154262199
                    0.996 0.343853154262199
                };
            \addplot [semithick, darkslategray63, forget plot]
            table {%
                    1.004 0.188207028341008
                    1.396 0.188207028341008
                };
            \addplot [semithick, darkslategray63, forget plot]
            table {%
                    1.604 0.37708216929552
                    1.996 0.37708216929552
                };
            \addplot [semithick, darkslategray63, forget plot]
            table {%
                    2.004 0.0924072183825149
                    2.396 0.0924072183825149
                };
        \end{axis}

    \end{tikzpicture}
    \caption{Lateral error separated by feature type. The observed error strongly depends on the map feature type, mostly due to different accuracy of the available road border features. The total lateral error is shown in blue and the offset corrected error in orange. We do not differentiate the number of drives in this plot. }
    \label{fig:boxplot_types}
    \vspace{-2mm}
\end{figure}
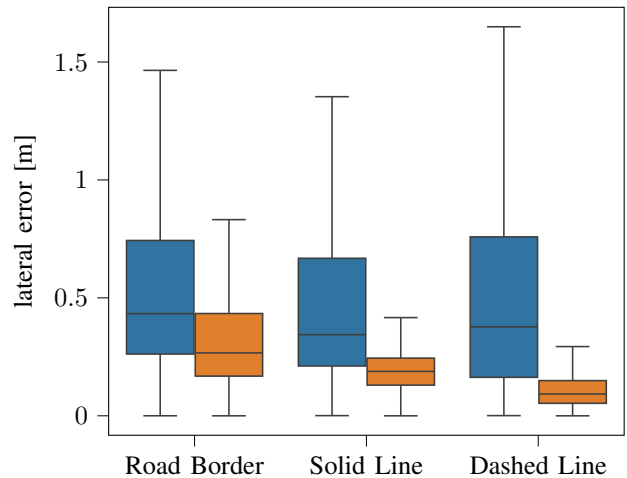

The evaluation metric is limited to the error in lateral driving direction, since longitudinal features are sparse in the measurement data.
We evaluate the lateral optimization result as described in section \ref{sec:expectation_maximization}.
The lateral error consists of two parts, the lateral \textit{offset} error resulting from an uncompensated localization error and the remaining \textit{non-offset} error, which is less correlated to the number of drives. This effect can be observed in Fig. \ref{fig:boxplot_number_of_drives_binned}.
The non-offset error is derived by first by computing an offset $o$ applied to all mapped points simultaneously, that minimizes the total error compared to the ground truth position. The remaining error is denoted as \textit{offset corrected error} and visualized in orange in Fig. \ref{fig:boxplot_number_of_drives_binned} and Fig. \ref{fig:boxplot_types}, contrary to the blue total error. In a real-world application of the method, this ground truth based offset correction is not possible, thus the number of drives visiting the mapped position should be kept well above 18 drives.

\subsection{Evaluation Results}
\label{sec:EvaluationResults}

The lateral accuracy this method achieves in best case scenarios with a high number of traversing drives and simple highway structures results in \textit{mean} total errors below $0.30$ meters.
Averaging all evaluated positions, including mapped exits and entry ramps and all map feature types results in a \textit{mean} total error of $0.49$ meters.
The mean absolute offset error as described in Fig. \ref{fig:evaluation_metrics} constitutes $0.41$ meters.
Correcting for this offset error, a resulting $0.27$ \textit{mean} error over all evaluated positions can be achieved in this theoretical setting.
For an increasing number of drives the offset corrected error and actual error converge as seen in Fig. \ref{fig:boxplot_number_of_drives_binned}.

Since we have multiple routes crossing the evaluated highway section, we have a varying number of drives at different points along the reference line.
Fig. \ref{fig:error_over_meters} illustrates this by showing the number of drives in comparison with the lateral error along the evaluated distance.
The relationship of the drive count at a mapped position with the lateral error is further investigated in Fig. \ref{fig:boxplot_number_of_drives_binned}.
As the optimization threshold criterion for valid results is not achieved at all times, eg. when passing some highway exits and ramps, the evaluation is only possible for around 90 percent of the ground truth reference line distance.

\begin{figure}
    \centering
    \begin{tikzpicture}

        \definecolor{darkgray176}{RGB}{176,176,176}
        \definecolor{darkslategray63}{RGB}{63,63,63}
        \definecolor{lightgray204}{RGB}{204,204,204}
        \definecolor{peru22412844}{RGB}{224,128,44}
        \definecolor{steelblue49115161}{RGB}{49,115,161}

        \begin{axis}[
                legend cell align={left},
                legend style={fill opacity=0.8, draw opacity=1, text opacity=1, draw=lightgray204},
                tick align=outside,
                tick pos=left,
                x grid style={darkgray176},
                xlabel={Binned Number of Drives},
                xmin=-0.5, xmax=6.5,
                xtick style={color=black},
                xtick={0,1,2,3,4,5,6},
                xticklabels={3+,6+,9+,12+,15+,18+,21+},
                y grid style={darkgray176},
                ylabel={lateral error [m]},
                ymin=-0.104065020424102, ymax=2.18536542890615,
                ytick style={color=black}
            ]
            \path [draw=darkslategray63, fill=steelblue49115161, semithick]
            (axis cs:-0.396,0.536285137046974)
            --(axis cs:-0.004,0.536285137046974)
            --(axis cs:-0.004,1.16656481935248)
            --(axis cs:-0.396,1.16656481935248)
            --(axis cs:-0.396,0.536285137046974)
            --cycle;
            \path [draw=darkslategray63, fill=peru22412844, semithick]
            (axis cs:0.004,0.121938017589598)
            --(axis cs:0.396,0.121938017589598)
            --(axis cs:0.396,0.318005598403208)
            --(axis cs:0.004,0.318005598403208)
            --(axis cs:0.004,0.121938017589598)
            --cycle;
            \path [draw=darkslategray63, fill=steelblue49115161, semithick]
            (axis cs:0.604,0.37122747024329)
            --(axis cs:0.996,0.37122747024329)
            --(axis cs:0.996,0.966792225363909)
            --(axis cs:0.604,0.966792225363909)
            --(axis cs:0.604,0.37122747024329)
            --cycle;
            \path [draw=darkslategray63, fill=peru22412844, semithick]
            (axis cs:1.004,0.0904187406273334)
            --(axis cs:1.396,0.0904187406273334)
            --(axis cs:1.396,0.318338823647474)
            --(axis cs:1.004,0.318338823647474)
            --(axis cs:1.004,0.0904187406273334)
            --cycle;
            \path [draw=darkslategray63, fill=steelblue49115161, semithick]
            (axis cs:1.604,0.223721136865732)
            --(axis cs:1.996,0.223721136865732)
            --(axis cs:1.996,0.612811152197822)
            --(axis cs:1.604,0.612811152197822)
            --(axis cs:1.604,0.223721136865732)
            --cycle;
            \path [draw=darkslategray63, fill=peru22412844, semithick]
            (axis cs:2.004,0.0819169763742098)
            --(axis cs:2.396,0.0819169763742098)
            --(axis cs:2.396,0.298736212001006)
            --(axis cs:2.004,0.298736212001006)
            --(axis cs:2.004,0.0819169763742098)
            --cycle;
            \path [draw=darkslategray63, fill=steelblue49115161, semithick]
            (axis cs:2.604,0.212694761209522)
            --(axis cs:2.996,0.212694761209522)
            --(axis cs:2.996,0.800534342111554)
            --(axis cs:2.604,0.800534342111554)
            --(axis cs:2.604,0.212694761209522)
            --cycle;
            \path [draw=darkslategray63, fill=peru22412844, semithick]
            (axis cs:3.004,0.0961453148155887)
            --(axis cs:3.396,0.0961453148155887)
            --(axis cs:3.396,0.2626374024484)
            --(axis cs:3.004,0.2626374024484)
            --(axis cs:3.004,0.0961453148155887)
            --cycle;
            \path [draw=darkslategray63, fill=steelblue49115161, semithick]
            (axis cs:3.604,0.190048763178327)
            --(axis cs:3.996,0.190048763178327)
            --(axis cs:3.996,0.57570890961682)
            --(axis cs:3.604,0.57570890961682)
            --(axis cs:3.604,0.190048763178327)
            --cycle;
            \path [draw=darkslategray63, fill=peru22412844, semithick]
            (axis cs:4.004,0.0938433702402497)
            --(axis cs:4.396,0.0938433702402497)
            --(axis cs:4.396,0.259252337126564)
            --(axis cs:4.004,0.259252337126564)
            --(axis cs:4.004,0.0938433702402497)
            --cycle;
            \path [draw=darkslategray63, fill=steelblue49115161, semithick]
            (axis cs:4.604,0.128064434458842)
            --(axis cs:4.996,0.128064434458842)
            --(axis cs:4.996,0.356775549040277)
            --(axis cs:4.604,0.356775549040277)
            --(axis cs:4.604,0.128064434458842)
            --cycle;
            \path [draw=darkslategray63, fill=peru22412844, semithick]
            (axis cs:5.004,0.0778443149024606)
            --(axis cs:5.396,0.0778443149024606)
            --(axis cs:5.396,0.25491417561462)
            --(axis cs:5.004,0.25491417561462)
            --(axis cs:5.004,0.0778443149024606)
            --cycle;
            \path [draw=darkslategray63, fill=steelblue49115161, semithick]
            (axis cs:5.604,0.159911216802004)
            --(axis cs:5.996,0.159911216802004)
            --(axis cs:5.996,0.348255958751902)
            --(axis cs:5.604,0.348255958751902)
            --(axis cs:5.604,0.159911216802004)
            --cycle;
            \path [draw=darkslategray63, fill=peru22412844, semithick]
            (axis cs:6.004,0.0996661246926739)
            --(axis cs:6.396,0.0996661246926739)
            --(axis cs:6.396,0.232734085996273)
            --(axis cs:6.004,0.232734085996273)
            --(axis cs:6.004,0.0996661246926739)
            --cycle;
            \draw[draw=darkslategray63,fill=steelblue49115161,line width=0.3pt] (axis cs:0,0) rectangle (axis cs:0,0);
            \addlegendimage{ybar,ybar legend,draw=darkslategray63,fill=steelblue49115161,line width=0.3pt}

            \draw[draw=darkslategray63,fill=peru22412844,line width=0.3pt] (axis cs:0,0) rectangle (axis cs:0,0);
            \addlegendimage{ybar,ybar legend,draw=darkslategray63,fill=peru22412844,line width=0.3pt}

            \addplot [semithick, darkslategray63, forget plot]
            table {%
                    -0.2 0.536285137046974
                    -0.2 0.00581169508715053
                };
            \addplot [semithick, darkslategray63, forget plot]
            table {%
                    -0.2 1.16656481935248
                    -0.2 2.08130040848204
                };
            \addplot [semithick, darkslategray63, forget plot]
            table {%
                    -0.298 0.00581169508715053
                    -0.102 0.00581169508715053
                };
            \addplot [semithick, darkslategray63, forget plot]
            table {%
                    -0.298 2.08130040848204
                    -0.102 2.08130040848204
                };
            \addplot [semithick, darkslategray63, forget plot]
            table {%
                    0.2 0.121938017589598
                    0.2 5.82076609134674e-11
                };
            \addplot [semithick, darkslategray63, forget plot]
            table {%
                    0.2 0.318005598403208
                    0.2 0.600512820061738
                };
            \addplot [semithick, darkslategray63, forget plot]
            table {%
                    0.102 5.82076609134674e-11
                    0.298 5.82076609134674e-11
                };
            \addplot [semithick, darkslategray63, forget plot]
            table {%
                    0.102 0.600512820061738
                    0.298 0.600512820061738
                };
            \addplot [semithick, darkslategray63, forget plot]
            table {%
                    0.8 0.37122747024329
                    0.8 0.00138750698622863
                };
            \addplot [semithick, darkslategray63, forget plot]
            table {%
                    0.8 0.966792225363909
                    0.8 1.85362854387168
                };
            \addplot [semithick, darkslategray63, forget plot]
            table {%
                    0.702 0.00138750698622863
                    0.898 0.00138750698622863
                };
            \addplot [semithick, darkslategray63, forget plot]
            table {%
                    0.702 1.85362854387168
                    0.898 1.85362854387168
                };
            \addplot [semithick, darkslategray63, forget plot]
            table {%
                    1.2 0.0904187406273334
                    1.2 0
                };
            \addplot [semithick, darkslategray63, forget plot]
            table {%
                    1.2 0.318338823647474
                    1.2 0.660005794014515
                };
            \addplot [semithick, darkslategray63, forget plot]
            table {%
                    1.102 0
                    1.298 0
                };
            \addplot [semithick, darkslategray63, forget plot]
            table {%
                    1.102 0.660005794014515
                    1.298 0.660005794014515
                };
            \addplot [semithick, darkslategray63, forget plot]
            table {%
                    1.8 0.223721136865732
                    1.8 0.00151645986756532
                };
            \addplot [semithick, darkslategray63, forget plot]
            table {%
                    1.8 0.612811152197822
                    1.8 1.19531829553467
                };
            \addplot [semithick, darkslategray63, forget plot]
            table {%
                    1.702 0.00151645986756532
                    1.898 0.00151645986756532
                };
            \addplot [semithick, darkslategray63, forget plot]
            table {%
                    1.702 1.19531829553467
                    1.898 1.19531829553467
                };
            \addplot [semithick, darkslategray63, forget plot]
            table {%
                    2.2 0.0819169763742098
                    2.2 0
                };
            \addplot [semithick, darkslategray63, forget plot]
            table {%
                    2.2 0.298736212001006
                    2.2 0.621038935083942
                };
            \addplot [semithick, darkslategray63, forget plot]
            table {%
                    2.102 0
                    2.298 0
                };
            \addplot [semithick, darkslategray63, forget plot]
            table {%
                    2.102 0.621038935083942
                    2.298 0.621038935083942
                };
            \addplot [semithick, darkslategray63, forget plot]
            table {%
                    2.8 0.212694761209522
                    2.8 0.00173018597646952
                };
            \addplot [semithick, darkslategray63, forget plot]
            table {%
                    2.8 0.800534342111554
                    2.8 1.68114613180545
                };
            \addplot [semithick, darkslategray63, forget plot]
            table {%
                    2.702 0.00173018597646952
                    2.898 0.00173018597646952
                };
            \addplot [semithick, darkslategray63, forget plot]
            table {%
                    2.702 1.68114613180545
                    2.898 1.68114613180545
                };
            \addplot [semithick, darkslategray63, forget plot]
            table {%
                    3.2 0.0961453148155887
                    3.2 5.82076609134674e-11
                };
            \addplot [semithick, darkslategray63, forget plot]
            table {%
                    3.2 0.2626374024484
                    3.2 0.511913596620363
                };
            \addplot [semithick, darkslategray63, forget plot]
            table {%
                    3.102 5.82076609134674e-11
                    3.298 5.82076609134674e-11
                };
            \addplot [semithick, darkslategray63, forget plot]
            table {%
                    3.102 0.511913596620363
                    3.298 0.511913596620363
                };
            \addplot [semithick, darkslategray63, forget plot]
            table {%
                    3.8 0.190048763178327
                    3.8 0.000151856014134849
                };
            \addplot [semithick, darkslategray63, forget plot]
            table {%
                    3.8 0.57570890961682
                    3.8 1.13644193989303
                };
            \addplot [semithick, darkslategray63, forget plot]
            table {%
                    3.702 0.000151856014134849
                    3.898 0.000151856014134849
                };
            \addplot [semithick, darkslategray63, forget plot]
            table {%
                    3.702 1.13644193989303
                    3.898 1.13644193989303
                };
            \addplot [semithick, darkslategray63, forget plot]
            table {%
                    4.2 0.0938433702402497
                    4.2 1.16415321826935e-09
                };
            \addplot [semithick, darkslategray63, forget plot]
            table {%
                    4.2 0.259252337126564
                    4.2 0.504558910857847
                };
            \addplot [semithick, darkslategray63, forget plot]
            table {%
                    4.102 1.16415321826935e-09
                    4.298 1.16415321826935e-09
                };
            \addplot [semithick, darkslategray63, forget plot]
            table {%
                    4.102 0.504558910857847
                    4.298 0.504558910857847
                };
            \addplot [semithick, darkslategray63, forget plot]
            table {%
                    4.8 0.128064434458842
                    4.8 0.000815781208947452
                };
            \addplot [semithick, darkslategray63, forget plot]
            table {%
                    4.8 0.356775549040277
                    4.8 0.695650383757382
                };
            \addplot [semithick, darkslategray63, forget plot]
            table {%
                    4.702 0.000815781208947452
                    4.898 0.000815781208947452
                };
            \addplot [semithick, darkslategray63, forget plot]
            table {%
                    4.702 0.695650383757382
                    4.898 0.695650383757382
                };
            \addplot [semithick, darkslategray63, forget plot]
            table {%
                    5.2 0.0778443149024606
                    5.2 0
                };
            \addplot [semithick, darkslategray63, forget plot]
            table {%
                    5.2 0.25491417561462
                    5.2 0.520080149597643
                };
            \addplot [semithick, darkslategray63, forget plot]
            table {%
                    5.102 0
                    5.298 0
                };
            \addplot [semithick, darkslategray63, forget plot]
            table {%
                    5.102 0.520080149597643
                    5.298 0.520080149597643
                };
            \addplot [semithick, darkslategray63, forget plot]
            table {%
                    5.8 0.159911216802004
                    5.8 0.000856396362733094
                };
            \addplot [semithick, darkslategray63, forget plot]
            table {%
                    5.8 0.348255958751902
                    5.8 0.628763503227492
                };
            \addplot [semithick, darkslategray63, forget plot]
            table {%
                    5.702 0.000856396362733094
                    5.898 0.000856396362733094
                };
            \addplot [semithick, darkslategray63, forget plot]
            table {%
                    5.702 0.628763503227492
                    5.898 0.628763503227492
                };
            \addplot [semithick, darkslategray63, forget plot]
            table {%
                    6.2 0.0996661246926739
                    6.2 0.00012743736988078
                };
            \addplot [semithick, darkslategray63, forget plot]
            table {%
                    6.2 0.232734085996273
                    6.2 0.42861706576627
                };
            \addplot [semithick, darkslategray63, forget plot]
            table {%
                    6.102 0.00012743736988078
                    6.298 0.00012743736988078
                };
            \addplot [semithick, darkslategray63, forget plot]
            table {%
                    6.102 0.42861706576627
                    6.298 0.42861706576627
                };
            \addplot [semithick, darkslategray63, forget plot]
            table {%
                    -0.396 0.818729287704229
                    -0.004 0.818729287704229
                };
            \addplot [semithick, darkslategray63, forget plot]
            table {%
                    0.004 0.203697595716402
                    0.396 0.203697595716402
                };
            \addplot [semithick, darkslategray63, forget plot]
            table {%
                    0.604 0.679768113190852
                    0.996 0.679768113190852
                };
            \addplot [semithick, darkslategray63, forget plot]
            table {%
                    1.004 0.182518745391618
                    1.396 0.182518745391618
                };
            \addplot [semithick, darkslategray63, forget plot]
            table {%
                    1.604 0.376804563591231
                    1.996 0.376804563591231
                };
            \addplot [semithick, darkslategray63, forget plot]
            table {%
                    2.004 0.172168980796935
                    2.396 0.172168980796935
                };
            \addplot [semithick, darkslategray63, forget plot]
            table {%
                    2.604 0.41127576072339
                    2.996 0.41127576072339
                };
            \addplot [semithick, darkslategray63, forget plot]
            table {%
                    3.004 0.162708326361599
                    3.396 0.162708326361599
                };
            \addplot [semithick, darkslategray63, forget plot]
            table {%
                    3.604 0.327800859520953
                    3.996 0.327800859520953
                };
            \addplot [semithick, darkslategray63, forget plot]
            table {%
                    4.004 0.161603564919936
                    4.396 0.161603564919936
                };
            \addplot [semithick, darkslategray63, forget plot]
            table {%
                    4.604 0.218512497857321
                    4.996 0.218512497857321
                };
            \addplot [semithick, darkslategray63, forget plot]
            table {%
                    5.004 0.160790688792744
                    5.396 0.160790688792744
                };
            \addplot [semithick, darkslategray63, forget plot]
            table {%
                    5.604 0.244970318071026
                    5.996 0.244970318071026
                };
            \addplot [semithick, darkslategray63, forget plot]
            table {%
                    6.004 0.175997138533756
                    6.396 0.175997138533756
                };
        \end{axis}

    \end{tikzpicture}
    \caption{Statistical box plot with lateral error of a position over the traversing number of drives at that position. The total lateral error is displayed in blue and the offset corrected error in orange. The error induced by localization offsets can be reduced with increasing number of drives. All box plots end whiskers at 1.5 times the interquartile range.}
    \label{fig:boxplot_number_of_drives_binned}
    \vspace{-2mm}
\end{figure}
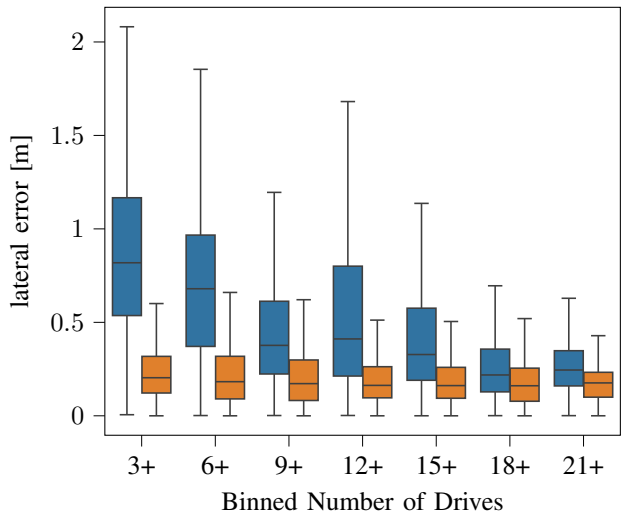

\section{Conclusions}
\label{sec:conclusion}

In this work, we proposed a system for a HD mapping pipeline for vehicle fleet data with processed sensor detections, making the approach scalable for large fleets.
The system is tolerant to missing or misclassified detections and can handle drives with multiple routes, generating a single complete map, model-free and without prior reference lines.

The pipeline uses randomly selected drives as pivot drives, along which a step-wise lateral sampling of detections is performed. These sampled points are then clustered and aligned using k-means and EM, estimating a lateral offset for each drive to compensate localization errors. The aggregated points are replaced with the maxima of their PDF and connected to form a polyline using a modified rectangular linear assignment algorithm. The data of different connected segments, e.g. from vehicles on varying routes, is then integrated into a hierarchical singular map graph.

The proposed approach achieves a high accuracy below 0.5 meters compared to a hand annotated ground truth map, as well as correctly resolving lane splits and merges, proving the feasibility of the use of series fleet data for the generation of highway HD maps. %
For future work, we would like to improve the reliability of the algorithm in complex scenarios and include other features such as traffic signs to correct misalignments in the longitudinal direction. %
\section*{Acknowledgements}

We would like to thank the Mercedes-Benz AG for providing the vehicle fleet data used in this work.
\printbibliography

\end{document}